\documentclass{article}

\PassOptionsToPackage{numbers, compress}{natbib}
\usepackage[final]{ap_2020}

\usepackage[utf8]{inputenc} % allow utf-8 input
\usepackage[T1]{fontenc}    % use 8-bit T1 fonts
\usepackage{hyperref}       % hyperlinks
\usepackage{url}            % simple URL typesetting
\usepackage{booktabs}       % professional-quality tables
\usepackage{amsfonts}       % blackboard math symbols
\usepackage{nicefrac}       % compact symbols for 1/2, etc.
\usepackage{microtype}      % microtypography
\usepackage{xcolor}
\usepackage{color}
\usepackage{colortbl}
\usepackage[colorlinks]{}
\usepackage{wrapfig,lipsum,booktabs}
\usepackage{listings}
\usepackage{subcaption}
\usepackage{graphicx}
\usepackage{enumitem}

\definecolor{burntorange}{rgb}{0.8, 0.33, 0.0}
\definecolor{light-gray}{gray}{0.9}
\definecolor{richelectricblue}{rgb}{0.03, 0.57, 0.82}

\title{Athena 2.0: Discourse and User Modeling \\ in Open Domain Dialogue}

\author{
	Omkar Patil, Lena Reed,  Kevin K. Bowden, Juraj Juraska, Wen Cui, Vrindavan Harrison,    \\
	\textbf{Rishi Rajasekaran, Angela Ramirez, Cecilia Li, Eduardo Zamora, Phillip Lee,  } \\
	\textbf{Jeshwanth Bheemanpally, Rohan Pandey, Adwait Ratnaparkhi}, and \textbf{Marilyn Walker} \\
	University of California, Santa Cruz \\
	Santa Cruz, CA  \\
	\texttt{\{opatil, vharriso, jjuraska, wcui7, lireed, kkbowden, } \\
	\texttt{rrajasek, aramir62, yli331, ezamora9, jbheeman, plee35,}  \\ 
	\texttt{jbheeman, rspandey, adratnap, mawalker\}@ucsc.edu} \\
}

\begin{document}

\maketitle

\begin{abstract}
Conversational agents are consistently growing in popularity and many people interact with
them every day. 
While many conversational agents act as personal assistants,  they can have many 
different goals. Some are task-oriented, such as providing customer support for a bank or 
making a reservation. Others  are designed to be empathetic and  to form emotional connections with the user. 
The Alexa Prize Challenge aims to create a socialbot,
which allows the user to engage in coherent conversations, on a range of popular topics 
that will interest the user.  Here we describe Athena 2.0, UCSC's conversational agent for Amazon's Socialbot Grand Challenge 4. Athena 2.0 utilizes a novel knowledge-grounded
discourse model that tracks the entity links that Athena  introduces into the dialogue, and uses them to constrain named-entity recognition and linking, and coreference resolution. Athena 2.0 also relies on a user model to personalize topic selection and other aspects of the conversation to individual users.

\end{abstract}

\section{Introduction}

Conversational agents are consistently growing in popularity and many people interact with
them every day. They help people
with day-to-day tasks, such as answering questions 
about the weather, creating shopping lists, and setting alarms. While many conversational agents act as personal assistants,  conversational agents can have many 
different goals. Some are task-oriented, such as providing customer support for finding movies
or making restaurants reservations \cite{mazumder2020application,shah2018bootstrapping,hakkani2016multi}. Others  are designed to be empathetic and  
to form emotional connections with the user \cite{zhou2020design,paranjape2020neural,finch2020emora,dinan2019build}. 
The Alexa Prize Competition aims to create a socialbot, that can converse coherently with
 users, on a range of popular topics 
that reflect users' interest \cite{gabriel2020further,fang2018sounding18,chen2018gunrock,bowden2019slugbot,curry2018alana}. Here we describe Athena, UCSC's conversational agent for Amazons Socialbot Grand Challenge 4.

A socialbot's goal is to engage the user by participating in a coherent conversation that the user enjoys. Athena  is capable of conversing on many different topics, and has a modular dialogue manager that supports  easy 
integration of new topics and response generators~\cite{harrison2020athena}.
In SGC4, we focused on  novel modules and experiments aimed at pushing forward the state of the art. These include: (1) developing a novel knowledge-grounded discourse model  that tracks the entity links that Athena  introduces into the dialogue, and uses them to constrain named-entity recognition and linking, and coreference resolution; (2) developing and comparing a rule-based and a neural
coreference engine; (3) improving the dialogue manager's neural response ranker; (4)  developing a user model and that modifies topic selection and response generation; (5) improving and increasing our  use of knowledge-graph response generators; and (6) developing several new neural response generators. Experiments with these
 was supported by Amazon's  new A/B testing regime, and our own  evaluation methods and analysis. 

\section{Architecture and System Overview }

\begin{figure*}
\includegraphics[width=5.5in]{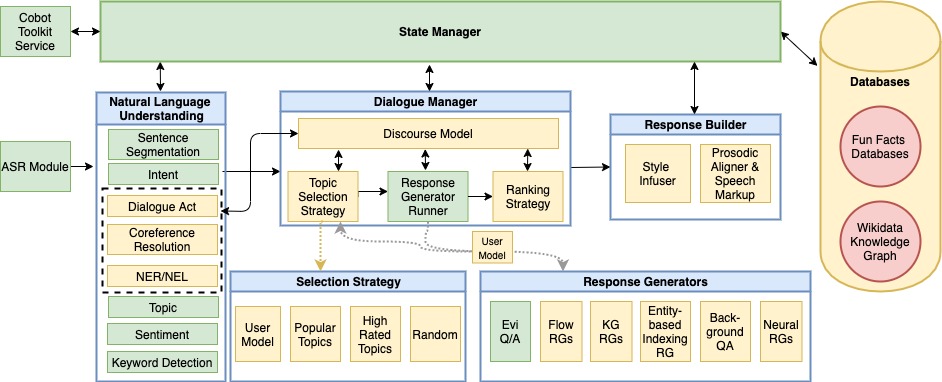}
  \caption{Athena's system architecture. Modules provided by Amazon are shown in green, and Athena's own modules are shown in yellow. 
 \label{fig:system-architecture}}
\end{figure*}

Figure~\ref{fig:system-architecture} details Athena's architecture. We built Athena using Amazon's Cobot Toolkit which depends on the Alexa Skills Kit (ASK)\footnote{\url{https://developer.amazon.com/en-US/alexa/alexa-skills-kit}}. Athena runs as an on-demand application that responds to ASK events containing utterance hypotheses produced by Amazon's automatic speech recognition (ASR) service.
Cobot also provides seamless integration with Amazon Web Services (AWS), and natively utilizes the 
AWS Lambda, DynamoDB, and ECS services ~\cite{khatri2018advancing}. 
We use several additional AWS services, such as the Neptune graph database, and Elastic Search.\footnote{\url{https://aws.amazon.com/elasticsearch-service/}}

The inputs to Athena are the ASR hypotheses for a user's turn, as well as a conversation ID that is used to retrieve the conversation history and state information from a back-end database. This is represented in the Discourse Model. See  Section ~\ref{dis-m-sec}. Then, the ASR hypothesis is fed into a natural language understanding (NLU) pipeline to produce a  representation of  the user utterance and conversation context (Section~\ref{nlu-sec}). Based on the NLU  and conversation context, the Dialogue Manager (DM) (Section~\ref{dia-m-sec}) updates the Discourse Model, and then dispatches a call to a set of response generators (RGs) to populate a response pool (Section~\ref{sec:response-generators}). The response generators that are called depend on the context and the NLU results. The Dial-M then applies a trained neural ranker to select Athena's next response from the potentially large pool of possible responses. This response is marked up with SSML \cite{taylor1997ssml}, and uttered using Amazon's text-to-speech (TTS) service.

The right hand side of Figure~\ref{fig:system-architecture} indicates how Athena uses the English subset of the  Wikidata Knowledge Graph,\footnote{\url{https://www.wikidata.org/wiki/Wikidata:Main\_Page}}  along with several fun facts databases organized by topic, named entity, and concept, and searched with Elastic Search. Large knowledge bases are essential for creating intelligent and versatile conversational agents \cite{fang2018sounding18,chen2018gunrock}. 
Athena's KG is constructed using Amazon's Neptune graph database, and accessed with SPARQL queries. Athena uses it for Named Entity Resolution (Section \ref{sec:ner}) and for the Knowledge-Graph Based RGs (KG RGs), as described in Section \ref{sec:kg-based-rg}.

\section{Natural Language Understanding }
\label{nlu-sec}
Athena's NLU  modules
consist of a combination of off-the-shelf modules
provide with Cobot, and Athena's own modules. See Figure~\ref{fig:system-architecture}.
We use Cobot's off-the-shelf tools for topic classification and intent recognition. We continued to use our own  
sentence segmentation,  and dialogue act tagging modules models from SGC3 (in green in Figure~\ref{fig:system-architecture}) \cite{harrison2020athena}.

This year our primary
focus for improving Athena's NLU was to develop a  discourse model. 
We invested considerable effort in improving Athena's named entity recognition and  linking modules, with novel training methods and data, and with the addition of constraints
from the discourse model. 
%The discourse model is also used to resolve anaphors to their entity links in Wikidata.

%In the following sections we describe Athena's individual components in greater detail.

\subsection{Discourse Model}
\label{dis-m-sec}

The Athena discourse model is designed to track the topics under discussion, selected named entities from the last few user and system utterances, speaker information, and anaphoric expressions. The number of entities to track is a parameter of the model, and is used to control the size of the ``centered'' entities \cite{BFP87,WIC90}. The information in the discourse model is then used in downstream tasks such as named entity recognition (NER), named entity linking (NEL), and system response generation. 

The discourse model facilitates information sharing by storing its data in a state table that is accessible to all Athena modules. This year's focus on the discourse model required changes to the Response Generators (RGs), which are responsible for creating the system response using a variety of strategies described in Section \ref{sec:response-generators}. 

Previously, the various Athena RGs did not share information in a well defined way, which led to overly short conversations and abrupt changes in topic. The current RG did not have direct knowledge of the entities introduced by previous RGs. This year, when the RGs introduce an entity in the system utterance, the entity and its knowledge graph ID are recorded in the discourse model. Then downstream RGs can then rely on the discourse model to identify any system generated entities and their IDs, rather than having the NER/NEL detect them in the system utterance.

%Furthermore, it is often necessary to share coreference information with downstream modules.  Since certain modules cannot deal with structured data and require strictly textual input, the discourse model applies a text ``augmenter'' to incorporate the output of the coreference step into the user utterance. The augmenter replaces pronoun spans with their corresponding entities, based on the output of the coreference step.

An important aspect of the discourse model is that it tracks if entities are introduced through a user utterance vs. a system response. This distinction is important for the confidence estimates given to the entities. The discourse model also contains information pertaining to the entity type (e.g. person, song, movie), which is used by  the coreference model to resolve pronouns. By introducing the discourse model and incorporating these different elements, Athena can produce conversations that focus on  entities that reflect the user's interest.

\subsection{Named Entity Recognition and Linking}
\label{sec:ner}
% Owner: Wen

%Named Entity Recognition (NER) and Named Entity Linking (NEL) play important roles in understanding entity mentions in users' utterances, and is critical for the functioning of Athena's Knowledge-Graph RGs, which retrieve facts from  Wikidata, HER EHERE providing entity information for downstream tasks, and allowing Athena's response generators to produce coherent responses. NER and NEL in open-domain conversations expose various challenges, such as the lack of annotated open-domain conversational data, the challenge of annotating a large number of real-user conversations when its not possible to use Mechanical Turk, due to the rules of the competition, and inaccurate recognition of popular entities with existing tools such as Spotlight DBPedia \cite{mendes2011dbpedia} and our Athena 1.0 NER/NEL module \cite{harrison2020athena}. To cope with these  challenges, we mined templates from real-user conversations, created $\sim$~1M instances of synthetic training data with context, leveraged a pre-trained BERT model \cite{devlin2018bert}, and developed a context-aware NER/NEL system. The NER/NEL in Athena 2.0  achieves a 15\%  improvement over the NER/NEL in Athena 1.0, and a  30\% improvement over Spotlight, when tested with Athena's real conversations.  

The named entity recognition (NER) and named entity linking (NEL) modules are used to understand entity mentions in users' utterances. They are critical for the Knowledge-Graph RGs, which use the entity information to retrieve facts from WikiData. NER and NEL also provide entity information for downstream tasks, and allow the response generators to produce coherent responses. NER and NEL in open domain conversations expose various challenges, such as the lack of annotated open-domain conversational data, and the inaccurate recognition of popular entities with existing tools such as Spotlight DBPedia \cite{mendes2011dbpedia} and our previous Athena 1.0 NER/NEL module \cite{harrison2020athena}. The creation of annotated data is especially difficult since the rules of the competition prohibit the use of Mechanical Turk on user data. To cope with these challenges, we mined templates from real-user conversations, created $\sim$~1M instances of synthetic training data with context, leveraged a pre-trained BERT model \cite{devlin2018bert}, and developed a context-aware NER/NEL system. The NER/NEL in Athena 2.0  achieves a 15\%  improvement over the NER/NEL in Athena 1.0, and a  30\% improvement over Spotlight, when tested with Athena's real conversations.

%\subsection{Popularity Metric}
% Owner: Phil, Wen

\noindent{\bf Popularity Metric.} We developed gazetteers by retrieving $\sim$950K Wikidata entities for the entity types needed to support  Athena’s popular topics such as Movies, TV, Music, Sports, Books, Board Games and Video Games. To help  generate the synthetic data, and provide features to the ranking algorithm, we also stored popularity scores based on English Wikipedia page views, aggregated over a three month period. We used Amazon's Elastic Search to create an index for these gazetteers for effective retrieval. 
%Gazetteers containing the most  popular entities can also be used to introduce entities in system initiatives when appropriate. 
The NER/NEL use popularity to disambiguate between two entities with the same name when there is no other distinguishing factor present. While the process of gathering page view counts is simple for most cases, niche genres required additional steps to generate a popularity score due to the way their Wikipedia and Wikidata pages are structured.   

%\noindent{\bf Synthetic Training Data Generation.} We sampled popular entities based on the popularity scores and gathered templates from real-user conversations, the team, and other datasets to generate synthetic data. The  templates from real-user conversations were 
%from  user responses where an entity was introduced by the 
% user, and the dialogue contexts are a sample of 5 to 15  contexts for the user response. Sample synthetic templates for book\_series and a sports\_team entities are shown in Table~\ref{table:synthetic_NER}. The template indicates the entity with a variable for its type. We then use the gazetteer to substitute the canonical name of the entity, as well as  its alternative names gathered from Wikidata, and the Wikidata ID for entity linking, as shown in the filled templates in Table~\ref{table:synthetic_NER}.
 
\noindent{\bf Synthetic Training Data Generation.} We create synthetic data by using templates derived from real user conversations and the Athena team. The templates are then instantiated with popular entities from WikiData. The  templates from real-user conversations were obtained from user responses where an entity was introduced in a user utterance. We also sample 5 to 15 dialogue contexts (previous system turns) for the user utterance. Sample synthetic templates for book\_series and a sports\_team entities are shown in Table~\ref{table:synthetic_NER}. The template indicates the entity with a variable for its type. We then substitute the canonical name of the entity, as well as its alternative names gathered from Wikidata, and the Wikidata ID for entity linking, as shown in the filled templates in Table~\ref{table:synthetic_NER}.

\begin{wraptable}{r}{3.4in}
    \vspace{-0.12in}
    %\begin{scriptsize}
    \small
     \begin{tabular}{@{} p{0.5in}|p{2.8in} @{}} \toprule
     Template & {\it Athena}: Have you read a series recently that you found really addictive?\\
        & {\it User}: i like [book\_series]\\ \hline
    Filled  &  {\it Athena}: Have you read a series recently that you found really addictive?\\
      Template  &  {\it User}: i like {\bf dresden files} (wiki id: Q2307373)\\ %\hline
    %Entity Used & wiki id: Q2307373 \\
            \hline
            \hline
    Template & {\it Athena}: Tell me a team you like.\\
        & {\it User}: i always like to watch [sports\_team] compete\\ \hline
    Filled  & {\it Athena}: Tell me a team you like.\\
    Template    & {\it User}: i always like to watch {\bf maple leafs}(wiki id: Q7826440) compete\\ \hline
    %Entity Used & wiki id: Q7826440 \\ \hline
    \end{tabular}
    %\end{scriptsize}
    \caption{ Synthetic Data Templates for Named-Entity Recognition and Linking  (NER/NEL) \label{table:synthetic_NER}}
    %\vspace{-0.15in}
\end{wraptable}
We organized the entities by popularity 
and only trained the model on the most popular instances. We held out entities for 
the test set. The topics and their entity types are movies, actors, directors, TV shows, songs, musicians, bands, athletes, sports teams, books, authors, board games, and video games.  For each entity type we generated approximately 100K training instances
and 10K test instances. Some entity types had fewer entity instances, such as sports teams, which resulted in fewer  generated instances.
%\begin{table}[ht]
\begin{wraptable}{r}{2.7in}
\centering.
    %\vspace{-0.15in}
    %\small
    \begin{tabular}{@{}llll@{}} \toprule
    {\bf Model} &  {\bf F1} & {\bf Prec.} & {\bf Recall}  \\ \hline \hline
    
    DBPedia Spotlight & 36\% & 78\%  & 23\%   \\ % wen
    NER/NEL SGC3 & 38\%  & 42\%  & 35\%  \\ % wen
    BERT Ranker Module & 47\%  & 54\%  & 41\%   \\%jesh
    % Current module & Synthetic & - & - & - \\ %jesh - no time to get
        \hline
    \end{tabular}
    %\vspace{-0.1in}
    \caption{ End-to-End NER/NEL results for a test set of  490 Athena user utterances with contexts. 
    \label{table:nel_result}} 
    %\end{table}
    \vspace{-0.1in}
    \end{wraptable}

%\subsection{BERT-based NER and NEL}
\noindent{\bf BERT-Based NER and NEL.} We first fine-tune a BERT model for mention-detection with the generated synthetic data, including dialogue context. For each mention, we retrieve candidate entities with their Wikidata IDs and popularity from the Elastic Search gazetteer, by fuzzy matching on the entity's canonical name and its alternative names. We employ another BERT model with cross-entropy loss to rank the candidates with features such as the dialogue context, popularity scores and similarity scores. We also create a test set of 490 real-user utterances from recent Athena conversations
% MAW> not sure what this means: that have richer entities 
to compare the new NER/NEL with  DBPedia Spotlight and the NER/NEL from SGC3. Table \ref{table:nel_result} shows that the new NER/NEL  improves both precision and recall by a significant magnitude. We are exploring further improvements using longer contexts from real user conversations, and different types of models.

\subsection{Entity-Based Coreference Resolution}
\label{sec:coref-resolution}
% Owner: Capstone Team

Coreference resolution relies on a discourse model. Once   pronouns are resolved to their antecedents, we can link them to entities in the KG, and store this information in the discourse model. This allows both further coreference resolution and named entity recognition and linking to access a more complete discourse model. We start with topics that we have KG RGs for (as described in Section~\ref{sec:kg-based-rg}), such as  music, tv shows, movies and sports because these RGs most
rely on entities from the  knowledge graph. From our logs, we are able to tell that the most frequent pronouns used include "he, she, him, her, they" and their derivatives, as well the pronoun "it". Therefore, our focus is currently only on resolution of these specific pronouns, rather than a general mentions resolution. We  also note that our approach is entity-based and doesn't consider antecedents that cannot be recognized by our NER/NEL such as event related referenced information like "watching movies" \cite{Byron02,Webber88a}. Figure~\ref{fig:convo_coref} shows conversational examples of coreference resolution.
%\begin{wrapfigure}{r}{3.1in}
\begin{figure*}[t!hb]
    %\vspace{-0.15in}
    %\begin{scriptsize}
    \small
     \begin{tabular}{@{} p{0.5in}|p{3.1in}|p{1.5in} @{}} \toprule
     {\bf Speaker} & {\bf Utterance} & {\bf Entities} \\
     \toprule
        ATHENA & {\bf The Office} is a great show. Why do you like {\bf it}? 	& The Office, it:The Office\\
        USER & {\bf It} is really funny &  It:The Office\\
        \hline
        \hline
        ATHENA & One of my favorite actors is {\bf Adam Driver}. Have you ever seen any of {\bf his} movies? 
        &Adam Driver, his:Adam Driver\\
        USER & Yes {\bf he} was in Star Wars &  he:Adam Driver \\
        \hline
        \hline
 	    ATHENA & {\bf Fleetwood Mac} has, so far, released 61 tracks. What do you think of that?
        &  Fleetwood Mac\\
    	USER & not as many as {\bf the beatles} but {\bf they} didn't write {\bf their} own stuff
    	& they:the beatles, \\
    	& & their: the beatles\\
    	\hline
    \end{tabular}
    %\end{scriptsize}
    \caption{Conversations with Pronominal Resolution    \label{fig:convo_coref}}
    %\vspace{-0.15in}
%\end{wrapfigure}
\end{figure*}

%\noindent{\bf Coreference Dataset}
\begin{wraptable}{r}{2.9in} 
%\begin{table}[h]
\vspace{-0.15in}
%\resizebox{2.9in}{!}{%
\begin{tabular}{@{}p{0.95in}|llll@{}}
\hline
  Anaphoric Form              & TV   & Music         & Sports     & Movie \\ \midrule
she, he, him, her  & 104  & 110           & 229       & 156   \\ 
it              & 812  & 751   & 85       & 1475   \\ 
they            & 0   & 30   & 0       & 47   \\ \bottomrule
\end{tabular}
\caption{Athena Conversation Data  Used for Testing the RB Coreference Model
\label{tab:coref-data}}
%\end{table}
\end{wraptable}
Unlike off-the-shelf coreference models which are usually trained on OntoNotes or CoNLL2012, we had the singular opportunity of using  large amounts of  data from Athena user conversations  to train the model to resolve coreference   in a dialogue setting. However, annotating coreference clusters is a tedious,  manual task. Currently,  3799 examples have been annotated by our team, see table~\ref{tab:coref-data} for details.  However, this is not enough for training a neural system. Therefore, we allocated a majority of these examples to our test set and augmented our training set with synthetic data. To create synthetic data, we gathered our annotated conversations, and grabbed common user responses to system utterances that we know are from handcrafted templates used by our knowledge based RGs to then slot fill for information. We then identify the slots that are in both the common user responses and system responses, and fill in accurate information using Wikidata. An example system and user template  is shown in Table~\ref{table:synthetic_coref} for the music topic.

\begin{wraptable}{r}{3.55in}
    %\vspace{-0.15in}
    %\begin{scriptsize}
    %\small
     \begin{tabular}{@{} p{0.55in}|p{2.8in} @{}} \toprule
     Template & {\it Athena}: Do you know [musician]'s songs from the year [year]? I really like [song] personally!\\	
        & {\it User}: yes I love [pronoun]\\ \hline
        Filled Template &  {\it Athena}: Do you know {\bf Ariana Grande}'s songs from the year {\bf 2019}? I really like {\bf thank u, next} personally!\\
      
        &  {\it User}: yes I love {\bf her}\\
        \hline
    \end{tabular}
    %\end{scriptsize}
    \caption{Synthetic Data Template and Sample Filled Template \label{table:synthetic_coref}}
    %\vspace{-0.15in}
\end{wraptable}

%Synthetic data is created by making templates from the most common user utterances and slot filling with the popular entities This resulted in around 500k synthetic data. 

%\subsection{Coreference Resolution Models}
To date, we have developed and experimented with several different models for coreference resolution: one rule-based model and two neural models. 
The benefits of a  rule-based system include  ease of implementation and efficiency. This approach is very lightweight compared to a neural model. An analysis of  data pulled from live traffic shows that the majority of anaphoric pronouns can be resolved with only one previous turn of context. This limits the scope and reduces the amount of context that needs to be processed. It is also easier to utilize the information within the discourse model with a rule- based system. Some contexts result in near 100\% confidence  in the prediction, and therefore can be captured with a small number of rules. The rule-based model uses the entities mentioned in the last utterance, as well as the type of the entity and the type of pronouns. The rule-based system outperforms off-the-shelf models such as Hugging Face's SpaCy neural coref \cite{clarkmanningdeep} and AllenNLP coref \cite{lee2018higher} on an annotated live traffic test set as shown in Table \ref{table:rules_model_scores}.
\begin{wraptable}{r}{8cm}
    \centering
    \begin{tabular}{ m{2cm} m{1.5cm} m{1.5cm} m{1.5cm} }
        \hline
         Model & F1 & Precision & Recall \\
         \hline
         Athena-RB & 69.93\% & 61.33\% & 81.33\% \\
         Allen AI & 25.8\% & 36.0\% & 20.19\% \\
         Hugging Face & 10.98\% & 12.98\% & 9.52\% \\
         \hline
    \end{tabular}
    \caption{Scores for Athena's Rule-Based (RB) model and baseline out-of-the-box Neural Coref Models.}
    \label{table:rules_model_scores}
\end{wraptable}

A drawback of a rule-based system is that it is heavily dependent on the quality of information that is present in the discourse model. However, conversation flows and user utterances can be unpredictable, and anaphora can be present in many different forms. For example, when pronouns refer to an entity within the user's current utterance, it causes complications because the entity has not yet been added to the discourse model (Figure \ref{fig:convo_coref} example 3). This is because the NLU pipeline runs coreference resolution before Named Entity Recognition and Linking (NER/NEL) for the current utterance, which allows NER/NEL to take advantage of the updated context provided by coreference engine. If  NER/NEL identifies the correct named entity, then that entity will be added to the discourse model.  However, neural models, such as AllenNLP's coreference model \cite{lee2018higher}, which  utilizes SpanBERT embeddings \cite{spanbert}, has a better chance at identifying these cases, as well as other problematic instances such as cataphora, and pleonastic pronouns. Therefore, we are exploring an ensemble model consisting of our rule-based system and a fine-tuned neural coreference model. 
%
% Dialogue Manager
%
\section{Dialogue Management}
\label{dia-m-sec}
\begin{wrapfigure}{r}{0.6\textwidth}
 %\vspace{-.2in}
% \begin{figure*}[hb]
  \centering
 \includegraphics[width=0.55\textwidth]{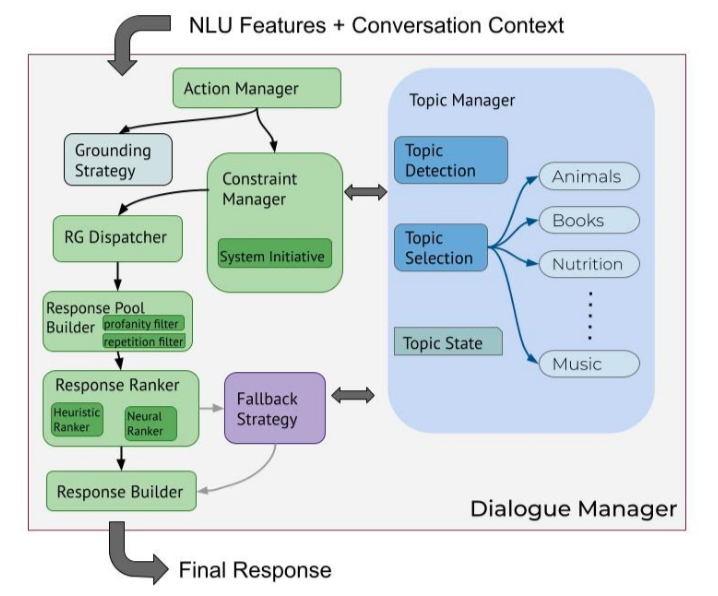}
  \caption{Athena 2.0 Dialogue Manager architecture.}
  \label{fig:dm-architecture}
  \vspace{-.2in}
% \end{figure*}
\end{wrapfigure}

\vspace{-.1in}
Open-domain conversational dialogue management is a challenging task due to the many different possibilities for valid responses to a user's utterance given  the context .  Unlike task-oriented dialogues, where the dialogue manager (DM) can optimize a clear set of objectives, open-domain dialogues do not have a clear metric of appropriateness \cite{lowe2017training,shalyminov2018neural,mehri-USR20,liu2016not,yietal19,higashinaka2019improving}. In order to tackle this problem, most successful Alexa Prize systems rely on using ``hand crafted'' scripted call-flows to form coherent dialogues on a particular topic. Athena’s DM uses a shallow hierarchy based on a top down decomposition into a number of subcomponents, each with its own responsibility. These  are oriented as a pipeline: the outputs of one component directly feed as inputs  to the next. The DM sub-modules are given in Figure~\ref{fig:dm-architecture}, and discussed below. 

In SGC4, we aimed to extend and improve Athena 1.0's  flexible and topic-agnostic approach to dialogue management. A good response should engage the user, and qualitatively be coherent in the context by  incorporating discourse salient  concepts and entities \cite{WPJ97,GJW95}. It is also important to   balance system and user initiative strategies  to give the users more control of the dialogue \cite{WW90,bowden2019cui}. 
We use many elements of the discourse model as input to the DM's trained response ranker, in order to maintain entity-based and topical coherence. This includes identifying topics of disinterest to user along with user confusion triggers like ``I don't know'' to make better decisions.

%\begin{table*}[h]
\begin{wrapfigure}{r}{2.8in}
\centering
\begin{small}
\vspace{-0.1in}
%\begin{tabular}{p{5.2in}} \toprule
\begin{tabular}{p{2.8in}} \toprule
 \texttt{ATHENA:} \textcolor{magenta}{Right.} 
 \textcolor{blue}{I would really like to go hiking in the woods.}
\textcolor{black}{What do you think is the most beautiful thing about nature?} \\ 
\midrule
 \texttt{USER:} when we go camping and the moons out and it's quiet
\\ \midrule
 \texttt{ATHENA:} \textcolor{magenta}{Ah, the wolves? Hmm.}
 \textcolor{blue}{Yes. I think we really need to preserve the beautiful nature around the world as much as we can.}
\textcolor{black}{There are some really beautiful national parks in the U.S.! 
Have you been to the Glacier National Park by any chance? It's in Montana.}
\\ \bottomrule
\end{tabular}
  \caption{Sub-conversation illustrating Athena's response parts construction. The {\sc ground} is magenta, the response {\sc opener} is blue, and the response {\sc body} is in black.}
  \label{table:response-parts-construction}
\end{small}
%\vspace{-0.2in}
%\end{table*}
\end{wrapfigure}

\noindent
\textbf{Response Structure.}
Athena's responses typically consist of  multiple speech acts. Athena’s responses can be decomposed into discrete parts that perform different dialogue purposes and are operated on separately. The subparts are called {\sc ground}, {\sc opener}, and {\sc body}. The example Athena response in Table~\ref{table:response-parts-construction} indicates each  subpart. 
The {\sc ground} always comes first:  it is limited to a strictly backward looking grounding dialogue act \cite{stolcke2000dialogue}. The {\sc opener} is dynamic and can be used for several purposes. It can be forward or backward looking.  The opener may be used to smooth the transition from one topic to another. It is commonly used to set up what will be said next by noting how it’s relevant: either directly to things previously said in conversation, or to the current topic. The response {\sc body}  contains new information that drives the conversation forward, and may contain multiple dialogue acts. The dialogue manager tracks the response parts separately.

\noindent
\textbf{Grounding Strategy.}
Given the response structure, every system utterance potentially includes a {\sc ground}
generated using either our prefixing functionality described in Section~\ref{sec:response-interweaving} or by individual Flow-RGs. 

\noindent
\textbf{RG-DM Interface.} 
The interface between the dialogue manager and response generators is a contract-based approach. The dialogue manager passes a set of response conditions to the response generators. The conditions act as a contract that the response generators must meet in order for their response to be considered for the current turn. This approach gives our dialogue system the flexibility to take advantage of a wide variety of response generator types, such as indexing and retrieval RGs, generative response models~\cite{harrison2019maximizing, oraby2019curate}, 
and call flows that resemble finite state automata. 
The DM-RG interface allows for backwards compatibility, e.g.  a previous version of Athena supported  a legacy system by treating it as an RG. In addition, completely swapping out all of Athena's RGs would require little modification to Athena's dialogue manager, consisting of updates to the response ranking function. 

\begin{wraptable}{r}{0.58\textwidth}
\vspace{-0.15in}
\def\arraystretch{0.8}
\begin{small}
    \begin{tabular}{p{0.9in}|p{1.9in} } \toprule
        \textbf{System action} & \textbf{Description} \\ \toprule 
        \texttt{perform\_repeat} & Repeat the previous turns utterance. \\ \midrule
        \texttt{conv\_closing} & End the conversation. \\ \midrule
        \texttt{advise\_usage} & Provide usage instructions. \\ \midrule
        \texttt{greet} & Start a conversation. \\ \midrule
        \texttt{repeat\_request} & Request that the user repeats themselves. \\ \midrule
        \texttt{wait\_prompting} & Wait for the user to think/finish what they were saying. \\ \midrule
        \texttt{red\_response} & Respond to a "red question". \\ \midrule
        \texttt{topic\_change} & Initiate a change of topic. \\ \midrule
        \texttt{list\_options} & Provide a verbal menu of options. \\ \midrule
        \texttt{converse} & Everything else. \\ 
         \bottomrule
    \end{tabular}
  \caption{Athena's system actions. }
  \label{table:dm-actions}
\end{small}
 \vspace{-0.1in}
\end{wraptable}
RGs can return multiple response candidates at each turn. In fact, this is encouraged in case some responses get removed during a quality filtering stage. If an RG does not have anything of substance to say at a given point in conversation, then it returns a null response (rather than saying something along the lines of ``I don't know what that is''). Furthermore, RGs are required to return the response parts as labeled separate segments.

\noindent
\textbf{Action Manager.}
Situations commonly arise where the dialogue system needs to facilitate the conversation by performing some functional speech act. These situations are topic independent and may occur at any point in dialogue. The job of the Action Manager is to recognize these situations and decide on an appropriate course of action for the system to take. We use the set of nine system action types in Table~\ref{table:dm-actions}. The \texttt{converse} action denotes the case where a functional speech act should not be performed. The Action manager uses a heuristic rule method based on NLU features, such as dialogue act, keyword matching, and regular expressions, to decide the next system action.

\noindent
\textbf{Constraint Manager.}
Once the system action has been decided, the Constraint Manager is responsible for generating a set of constraints that describe the next system utterance. The constraints specify a topic of discussion, as well as potential entities to mention. Also, the constraints contain a flag that signals if the dialogue manager is initiating discussion on a new topic this turn.  For example, the constraints: \texttt{\{ Topic = harry\_potter, Entity Mention = J.K\_Rowling,  Dialogue\_act = opinion\_question \}}, 
describe the following system utterance:
“\textit{What do you think about J.K. Rowling? Do you know much about her?}”

Currently, the constraint generation is based on heuristic rules. The constraint manager is designed using a flexible topic-agnostic approach  based on general principles of entity/topical coherence. In short, the general strategy of the constraint manager is to talk about the same entities or concepts as the user. The constraint manager can specify soft or hard constraints. 

% The dialogue act is a soft constraint, and we attempt to use a variety of dialogue act types to increase engagement. Primarily, the DM alternates between opinion statements, non-opinion statement, and opinion questions.    
\begin{wraptable}{r}{0.3\textwidth}
\vspace{-0.1in}
\begin{small}
    \begin{tabular}{ll}
        \toprule
        % \multicolumn{2}{c}{Topics} \\
        % \midrule
        \textit{sports} & \textit{movies} \\
        \textit{books} & \textit{nature} \\
        \textit{news} & \textit{animals} \\
        \textit{astronomy} & \textit{comic books} \\
        \textit{dinosaurs} & \textit{harry potter} \\
        \textit{nutrition} & \textit{pirates} \\
        \textit{video games} & \textit{board games} \\
         \textit{TV} & \textit{food} \\
        \textit{hobbies} & \\ %\textit{science and} \\
         %& \textit{technology} \\
        %\textit{introduction} & \textit{persona} \\
        %\textit{controversial} & \textit{politics} \\
         %&  \\
         \bottomrule
    \end{tabular}
  \caption{Athena's topics.}
  \label{table:athena-topics}
 \end{small}
 \vspace{-0.1in}
\end{wraptable}
In the process of generating constraints for the next turn, the Constraint Manager asks some internal questions: What is the current topic? Is it a continuation from the last turn's topic or has the user changed it to something new? Are we still on the same topic, but have shifted to a new sub-topic, or focused on a new entity? These questions are answered by another module: the Topic Manager sub-module, described below. 

While generating constraints, the DM has the ability to initiate a topic change action. Two of the most common reasons for changing the topic during this phase are if the DM loses track of the current topic, or does not know appropriate constraints to produce for current conversational context.
% Conversation constraint generation is one area that we are very interested in from a research perspective, and we are currently  constructing a machine learning model based approach to constraint generation.  

\noindent
\textbf{Topic Manager.}
%\textcolor{red}{Describe Topic manager by writing out information from the Amazon site visit presentation slides on topic manager.}
The Topic Manager refers to a collection of modules and sub-modules that perform several tasks related to topic representation, topic detection, and topic state representation.  Athena's dialogue manager uses a topic state to track topical information related to the current conversation so far. In addition to the current topic, the topic state includes a distribution over turns spent on each topic so far, and a topic history containing the sequence of topics visited. It contains lists of entities mentioned by both the user and the system. Only some of  Athena's RGs  supply responses with annotated entity mentions. Flow-RGs that have a KG RG also have a dictionary that can return Wikidata IDs for
any entity mentions, in order to support the Discourse Model and coreference across RGs.

Individual topics are represented using Python classes. Each class stores information related to that topic such as its name, referential expressions associated with the topic, and subtopics that fall within its general domain, e.g., basketball is a subtopic of sports. The arrangement of topics is designed to reflect Athena's conversation structures and content.
% needs rather than real world topical relationships. For instance, one might consider artificial intelligence a subtopic of science and technology rather than a separate topic, which is how they are represented within Athena.
Table~\ref{table:athena-topics} lists all the topics that Athena specifically has targeted RGs for. Other topics are handled by retrieval based methods, or by search, such as by using EVI, a database of facts stored in an Elastic Search index, and other response indices that are based on user's frequent utterances and offer a few turns of conversation on that topic. 

The topic detector module is also part of the Topic Manager. 
The topic detector searches the user utterance for topic name mentions and key words associated with each topic. Then, a combination of features, such as the  dialogue act, topic key-words, and Amazon' Topic classifier are used to recognize and detect situations where the user is trying to invoke some new topic. We categorize user topic invocations into two types. In the first type, the user explicitly invokes the topic by name, e.g., "let's talk about sports". In the second type, the user might only refer to some entity related to Athena's internal topic, e.g., "what do you think about Kobe Bryant?" falls within the sports topic class. Athena's  dialogue manager  requires  a discrete topic class to be assigned to each conversation turn. Therefore, correctly identifying entities and linking them to their associated topic is paramount. 

\noindent
\textbf{Initiative Manager.}
In some situations the DM may decide to change the topic of discussion. The Initiative Manager comes into play after the decision has been made to leave the current topic. It decides  what new topic to initiate, or whether to give the user the initiative. We have  repeatedly experimented with different approaches to allowing the user more initiative.

\begin{wrapfigure}{r}{3.5in}
%\begin{figure}
  \vspace{-.1in}
  \centering
  \includegraphics[width=3.6in]{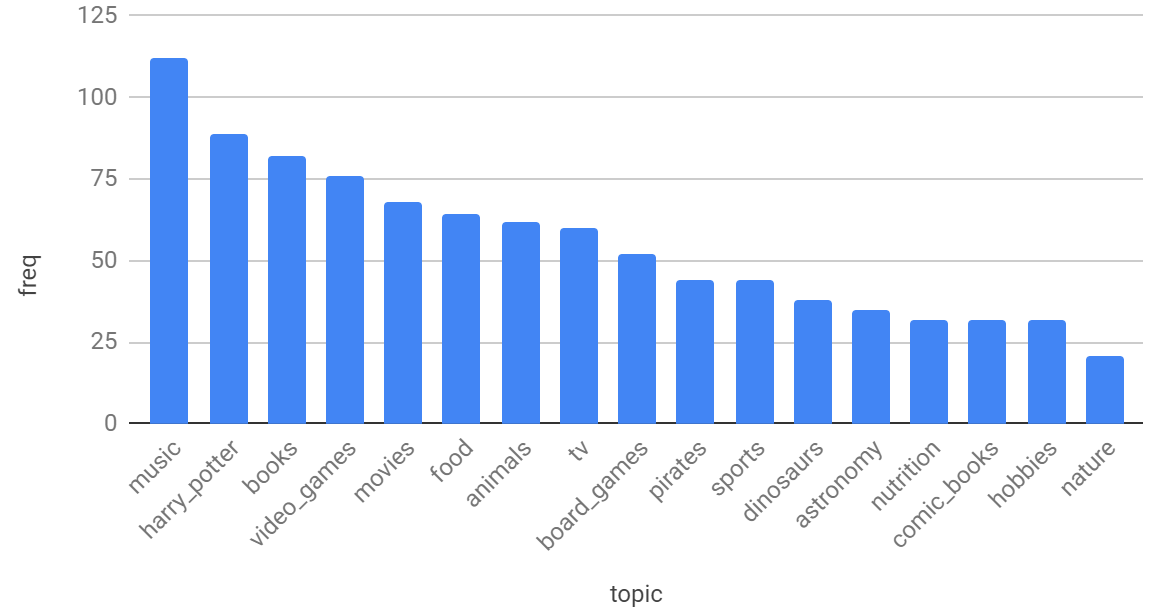}
  \vspace{-.2in}
  \caption{A distribution of the topics selected after we've given the user a menu of options.}
  \label{fig:give_user_init_dist}
  \vspace{-.1in}
\end{wrapfigure}

In SGC4, we increased the frequency of our menu-prompt that offers the user a selection of  three topics that haven't been discussed yet, to help the user take initiative in the conversation.  Figure~\ref{fig:give_user_init_dist} shows  the distribution of Athena topics selected as a result of a menu prompt. 

%\end{figure}

\noindent
\textbf{RG Dispatcher.}
The RG dispatcher decides which RGs to run given the action and constraints. In most cases Athena has two or three RGs that can contribute to conversations within a given topic.  All RGs capable of meeting the constraints are selected. Also,  some RGs run every turn, such as  RGs for responding to “red” questions, RGs that contain Athena’s persona responses, and question-answering RGs such as EVI. 

The decision making in the response dispatcher is an automatic process mainly achieved through look up tables. We construct a registry by requiring each RG to register (at time of instantiation) the action types and topics that they are capable of responding to. Then, during execution, the registry determines which RGs to run based on the outputs of the Action Manager and Constraint Manager.

\noindent
\textbf{Response Pool Builder.}
The response pool builder is the DM module that interfaces with the response generators. This module takes the list of RGs output by the RG Dispatcher and calls on them to populate a preliminary response pool. This preliminary response pool is refined with filters for profanity and repetition and then gets sent back to the DM. Before passing the response candidates to Cobot's offensive speech classifier, we mask words and phrases that we observed to cause responses to be flagged falsely as offensive, such as ``king'', ``president'', or ``saturated''.

\noindent
\textbf{Response Ranker.}
The goal of the response ranker is to find the best response that also matches the constraints output by the Constraint Manager. 
Athena's  response ranker  is  a  BERT-based neural response ranker which was trained on hand annotated Alexa Prize conversation data. The annotators were given two turns each of user and Athena's conversation context along with a pool of candidate responses, and asked to rank the candidate responses. We have  annotated additional data and retrained this response ranker three times during SGC4. 

\noindent
\textbf{Fallback Strategy.}
Several circumstances can lead to the response pool being empty. For instance, a system crash,  the RGs could not meet the dialogue manager constraints, or all the responses were removed from the response pool during the filtering stage. The purpose of the Fallback Strategy module is to take over and provide a graceful way of moving the conversation forward.  Our most frequent fallback approach is to initiate conversation on a new previously unvisited topic, but after several system initiatives, Athena will use a fallback strategy that gives the user the initiative by prompting for an unvisited topic.  The Fallback Strategy selects a template utterance from a collection of utterances crafted specially for this circumstance.  

\noindent
\textbf{Response Builder.} 
The Response Builder module assembles the response parts and performs a cleaning post-processing step, before sending the response to the TTS engine.

\section{Response Generation}
\label{sec:response-generators}

%\subsection{Response Generation Approaches}

We believe that having multiple response generators, of different types, per topic, is essential to dynamic creation of unique dialogue interactions, that inject stylistic variations within topic into the conversations. For additional details please see the Athena 1.0  Technical report for SGC3. ~\cite{harrison2020athena}.
We describe our approach to flow-based generation in Section~\ref{sec:flow-based-rg}, and describe two specific special cases of flow-RGs. In  Section~\ref{sec:intro_rg},
we describe our Introduction flow and in Section~\ref{sec:hyp-and-wyr-questions} we describe a general flow framework for topic specific Would-You-Rather and Hypothetical Questions.
Section~\ref{sec:kg-based-rg} then describes our Knowledge-Graph based RGs. Section~\ref{sec:centering-rg} describes our framework for Centering RG's, and Section~\ref{nrg-sec} describes our Neural RGs.

%\subsubsection{Evi (general question answering)}
%\subsubsection{Retrieval (Washington Post news, Reddit fun facts)}

\subsection{Flow-Based Response Generators}
\label{sec:flow-based-rg}
% Owner: Jurik

Though lacking somewhat in robustness, and requiring a great deal of handcrafting, call-flows are still the most reliable way to generate contextually appropriate responses in a dialogue system. With a graph of nodes between which the RG moves depending on what the user replies, and with a set of system response templates for each node to select from, flow-based RGs capture a larger and more specific context, and are thus able to be more  context-aware and fluent, when compared to  other RG types, such as retrieval-based or neural. This RG design naturally has a rather limited support for user initiative, which we make up for with other RGs in Athena and by ensuring the responses from different RGs get smoothly interwoven across multiple turns, as well as within a single turn.

Our Flow-RG framework was  created for Athena 1.0 to supports the design of dialogue flows and systematically communicate information from the Flow RGs to the DM. Flow-RG functions to modularize and  streamline the process of implementing flow-based RGs, and to impose a common structure and constraints on RGs across all topics. Each call-flow (typically one for a topic) can comprise multiple independent \emph{miniflows} (corresponding to subtopics or specific subjects), and Flow-RG ensures the transitions between them are smooth and natural, regardless of which order they are executed in. Since flow-based RGs tend to lean more on system initiative, the framework also ensures that interweaving with other types of RGs (such as for question answering), and the subsequent flow resumption, work seamlessly, as described and illustrated in Section~\ref{sec:response-interweaving}.

Flow-RG enforces each next turn in the flow graph to be conditioned on the dialogue act(s) of the current user utterance, which should reduce the chance of Athena's subsequent response ignoring the user's intent, which can be anything from expressing an opinion, to requesting information, to merely acknowledging Athena's response in the previous turn ~\cite{harrison2020athena}. To make the conditioning more robust, we now allow \emph{preconditions} based on other aspects of the user utterance to be specified as well, such as a named entity or a particular set of keywords present in the utterance, or the sentiment of the utterance. Another feature we introduced in Flow-RG this year are \emph{universal templated miniflows}, which only need a data file to be provided in order for them to be enabled for a given topic. Two examples of these are the hypothetical and would-you-rather question miniflows introduced in the majority of Athena's flow-based RGs (see Section~\ref{sec:hyp-and-wyr-questions}).

\begin{figure*}[tb]
%\begin{wraptable}{l}{3.1in}
    %\vspace{-0.15in}
    %\begin{scriptsize}
    \small
    %\begin{tabular}{@{} p{0.5in}|p{2.4in} @{}} \toprule
    \begin{tabular}{@{} p{0.5in}|p{4.7in} @{}} \toprule
        ATHENA & Okay, can't wait to get to know a thing or two about you. Valentine's day is on Sunday. Do you have a Valentine?\\
        USER &  yes my wife\\
        ATHENA & Cool! I wish I could meet a socialbot that I could fall in love with; then they'd be my Valentine. Anyway, I've always wanted to go on a vacation and to travel. What's a place that you've always wanted to travel to?\\
        USER & paris sounds great\\
        ATHENA & the eiffel tower is a masterpiece of architecture. For me, I think the great wall of china is at the top of my list of travel destinations. It seems like it would be amazing to walk along it. So, Is there a particular reason why you want to go to paris?\\
        USER & the food is really bomb\\
        ATHENA & Delicious food is always a good reason to go somewhere. Hey, I'm curious, Do you prefer flying? or road trips?\\
        USER & I like flying generally\\
        ATHENA & That's an interesting answer. I'd like to go on the road. You can see so much during a roadtrip, it would be cool to watch the world go by. Hmm, Do you like to relax during vacations?\\
        \hline
    \end{tabular}
    %\end{scriptsize}
    \caption{An excerpt from the Introduction Flow-RG, taking place right before Valentine's day. \label{fig:intro_example}}
    %\vspace{-0.15in}
\end{figure*}
%\end{wraptable}

In SGC4,  we take full advantage of the Flow-RG framework and adopt it this time for \emph{all} the topics that Athena supports. We further expand the capabilities of the framework, making it even easier for robust and dynamic flow-based RGs to be created, and for existing RGs to be extended. Although we add RGs for new topics, such as food, we focus primarily on extending the depth of conversations Athena can have on individual topics.  More details on the Flow-RG framework and its features can be found in our SGC3 report \cite{harrison2020athena}.

\subsubsection{Response Generator Interweaving}
\label{sec:response-interweaving}

\begin{figure}[htb]
    % \vspace{0.2cm}
    \includegraphics[width=\textwidth]{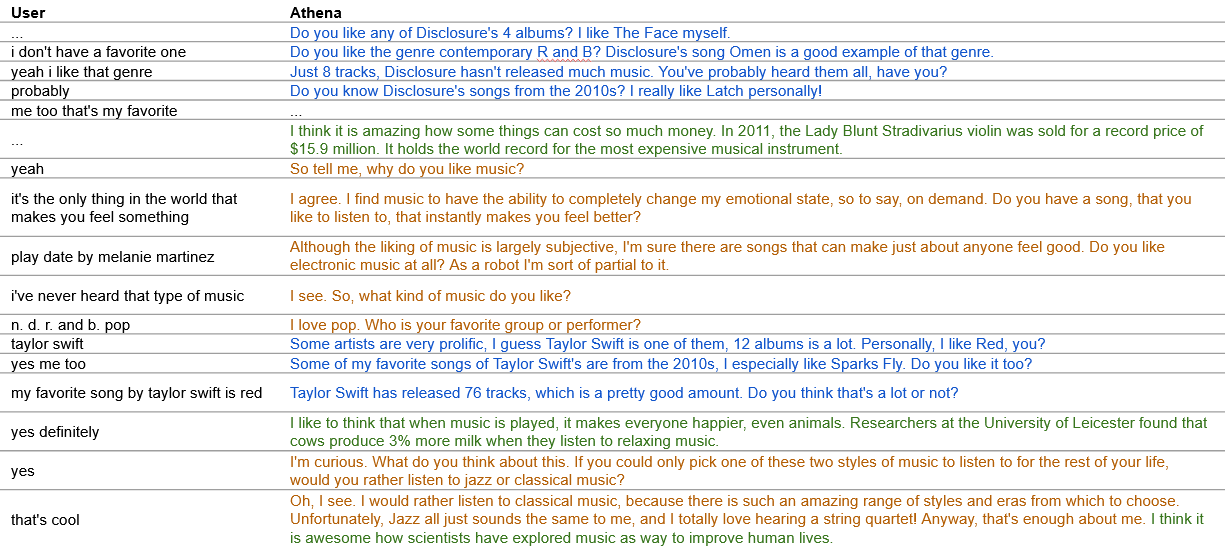}
    \caption{An excerpt of a conversation in which Athena discusses music with the user, interweaving the responses of 3 different RGs. Responses from the KG-based RG are highlighted in blue, responses from the Center RG (fun facts) in green, and those from the flow-based RG in orange.}
    \label{fig:rg-interweaving-example}
    \vspace{-0.1cm}
\end{figure}
The Flow-RG framework is responsible for Athena's unique capability for response interweaving, which guarantees that  Athena will never use the same sequence of dialogue acts in any two conversations.
We refine the multi-part response structure in Athena introduced last year, and use it to further improve the RG interweaving capabilities of our system. With 2-3 RGs for almost each topic in this year's system, and other non-topic-specific RGs that can offer a response at any point in the conversation, we had to make the interweaving mechanism more robust, so that the transitions between turns, as well as within a single turn, would remain smooth and natural. This is a fundamental part of our system's design, which relies on the response ranker to make the decision about the best next response in the current context from a pool of multiple candidates at each turn.

Figure~\ref{fig:rg-interweaving-example} demonstrates the interweaving behavior across a longer subdialogue about music, which is a topic supported by 3 distinct RGs in Athena. In the example, each of the RGs handles 1 to 5 turns at a time, contributing to overall more than 15 consecutive turns on a single topic, without the user expressing a desire to change the topic. There is only one instance of RG interweaving within the same turn in this particular example, and that in the last turn.

An extension of our response interweaving is our ability to prefix reactive utterances to the beginning of the next response. Figure ~\ref{fig:sb_concat_example} provides  two examples. These have multiple functions, such as answering the user's questions, grounding the user's input, or reacting to the user's feedback such as complaints or compliments. These prefixes are critical for addressing unexpected user input, and making the user feel like Athena is listening without losing the flow of the current topic.

 \begin{wrapfigure}{r}{3.1in}
    \vspace{-0.15in}
    %\begin{scriptsize}
    \small
    \begin{tabular}{@{} p{0.5in}|p{2.5in} @{}} \toprule
        ATHENA & Thanks for the information! You know, I realize I've been asking you a lot of questions, but how about me? Is there anything you want to know about me? \\
        USER & what is your favorite song\\
        ATHENA & \textcolor{burntorange}{I like a lot of different songs, but one of my favorites is, Piano Man, by Billy Joel. But anyways, }I appreciate you wanting to learn more about me, and I definitely want to keep chatting. I was wondering, do you think it's easier to break the ice with a human, or an AI like me? \\
        \hline
        \hline
        ATHENA & Spiders are so interesting. Here's a fact I read about them! There are over 100 species of spiders that mimic ants by having evolved similar appearances and even similar pheromones. Most do it to evade predators, but a few do it to help them prey on ants.  \\
        USER & that's so cool \\
        ATHENA & \textcolor{burntorange}{yeah, I find it interesting too. Hhmm, anyways, }So, what's your favorite animal? \\
        \hline
    \end{tabular}
    %\end{scriptsize}
    \caption{Two examples of our reactive-response prefixing, both highlighted in orange. The first example prefixes an answer to the user's question, while the second example prefixes an  acknowledgement.   \label{fig:sb_concat_example}}
    %\vspace{-0.15in}
\end{wrapfigure}

\subsubsection{Introduction RG}
\label{sec:intro_rg}
% Owner: Kevin

One specific Flow-RG that has been important to our system is our Introduction. Among all flow-based RGs, the one responsible for Athena's chit-chat at the beginning of each conversation was expanded the most substantially, since every user talking to Athena interacts with it, and we observed previously that the initial chit-chat can have a strong positive effect on the users' overall experience. 

We use our Introduction to front-load the conversation with getting to know you content. Here we ask icebreaker questions, such as learning their name, their bucket list travel destinations, and weekend activities. We also use this as an opportunity to bring up relevant current events, such as holidays, and discuss their work from home/vaccination situation. We also give the user a chance to ask us questions and give us advice. Figure~\ref{fig:intro_example} demonstrates some of these features. The introduction has been carefully crafted to get the conversation off on the right foot, and to learn information for the user model. The introduction also changes significantly for repeat users to not ask the same questions, and to mention aspects of the previous conversation.

\subsubsection{Would-You-Rather and Hypothetical Questions}
\label{sec:hyp-and-wyr-questions}
% Owner: Kevin

%\begin{wraptable}{l}{3.1in}
\begin{figure*}[th]
    %\vspace{-0.15in}
    %\begin{scriptsize}
    \small
  %  \begin{tabular}{@{} p{0.5in}|p{2.4in} @{}} \toprule
   \begin{tabular}{@{} p{0.5in}|p{4.7in} @{}} \toprule
        ATHENA & I'd like to hear your ideas on this. If you could choose one food to live on for your entire life, which food would it be? Why? \\
        USER & i think nachos \\
        ATHENA & That's an interesting answer! If I was a human, I would eat Sweet potatoes. Sweet potatoes are a really nutritious food. You can eat them in a variety of ways like baked, fries, or twice baked. Anyway, that's my thoughts on the matter. \\
        \hline
        \hline
        ATHENA & Okay. I was curious about your opinion on this. If you had the option, would you rather spend time in the mountains, or, at the beach? \\
        USER & i like the beach more \\
        ATHENA & Choosing the beach is a good choice! If I was a human, I would spend time at the ocean - I would walk on the beach, search for shells, and relax in the sun. Anyhow, that's where my minds at, let's move forward. \\
        \hline
    \end{tabular}
    %\end{scriptsize}
    \caption{An example of a Would-You-Rather sequence and Hypothetical sequence in the Food and Nature topics respectively. \label{fig:wyr_and_hq}}
    %\vspace{-0.15in}
\end{figure*}   
%\end{wraptable}

Athena is able to ask both a Would-You-Rather and Hypothetical question to extend our conversation on the current topic once per topic. An example of each question type can be seen in Figure~\ref{fig:wyr_and_hq}. Hypothetical questions have more open-ended responses and may lead to misunderstanding the user. These questions were crowdsourced and used in the SlugBot version of our system~\cite{bowden2019cui}. For SGC4, we filtered out lower quality questions, and edited the content to be inline with Athena's persona. In two separate live A/B experiments,  conversations using these questions had a statistically significant higher conversational length and user rating.

\subsection{Knowledge Graph-Based Generators}
\label{sec:kg-based-rg}
%Owner: Lena
The goal of the Knowledge Graph-based Response Generators (KG RGs) was to 
create deep knowledge-grounded conversations where you never run out of 
things to say on a topic.  Our KG RGs are created using Wikidata, which we migrated to this year after using
the Amazon Knowledge Graph previously. We
query a copy of the Wikidata database on Neptune using the query language SPARQL. We have
implemented four topics which access Wikidata entities and relations to form responses,
movies, music, sports and TV shows. The conversations are anchored around named 
entities. Each topic attempts to continue the conversation by 
either responding with a fact about the entity which was mentioned in the previous turn, 
an entity mentioned by the user, or an entity from a set of fallback entities. When 
the system has either run out of facts on a particular entity or has been on the same 
entity for a number of turns above a threshold, the system attempts to switch to a related
entity to continue the conversation. 

\begin{wrapfigure}{r}{2.7in}
    \vspace{-0.15in}
    %\begin{scriptsize}
    \small
    \begin{tabular}{@{} p{0.5in}|p{2.2in} @{}} \toprule
        ATHENA & Is there a television show that you like? \\
        USER & {\bf Friends} is one of my favorites \\
        & Entity linking: Friends -> entity:Friends, wiki\_id: Q79784\\
        ATHENA & Friends is a great show. Why do you like it? \\
        USER & It is really funny \\
        ATHENA & What character do you like the most? \\
        USER & I love Rachel \\
        ATHENA & I really like how Jennifer Aniston plays Rachel 
        Green. Did you know that she won a Primetime Emmy Award
        for Outstanding Lead Actress in a Comedy Series for her
        role? \\
        USER & no, that's interesting \\
        \hline
    \end{tabular}
    %\end{scriptsize}
    \caption{Conversation using the TV KG      \label{fig:convo_kg}}
    \vspace{-0.15in}
\end{wrapfigure}
Each topic has one to three entity types that the RG can respond about, and  each
entity has a set of relations used to generate 
responses. The entity types and relations for  each topic 
are in Table~\ref{table:kg_topics}. 
Each relation can only be used once for a particular entity, but can be reused
when the RG has switched to a new entity. 

With the TV show RG, we implemented a more dialogic KG RG. Previously, the data 
gathered from each relation was presented as a fact with a follow-up question
prompting a user response appended to the end. Our updates allow us to also 
include turns which start with a question and then use the KG data to respond
based on the user's input, an example of which is in Figure~\ref{fig:convo_kg},
We are also able to generate 
other multi-turn responses, such as asking if the user knows a fact about a 
related entity before presenting that information. 
We additionally updated our responses based on information gathered from the user
which indicates if they are familiar with the show we are currently discussing.
This allows us to ask questions that depend on the user having seen the show, which
we could not ask previously.

\begin{wraptable}{r}{3.4in}
    \vspace{-0.15in}
    %\begin{scriptsize}
    \small
    \begin{tabular}{@{} p{0.5in}|p{0.8in}|p{1.6in} @{}} \toprule
        \bf Topic & \bf Focus Entities & \bf Relations \\ \hline
        MOVIES & Movies, Actors, Directors & cast, voiceCast, 
        spouse, childrenNum, genre, awards \\ \hline
        MUSIC & Musicians, Bands & performer (song and album), 
        genre, awards, memberOf, instrument, label \\ \hline
        SPORTS & Athletes, Sports & team, position, participant 
        (tournament, leagues), spouse, childrenNum, awards \\ \hline
        TV SHOWS & TV Shows & cast, role, creator, director, genre, awards \\
        \hline
    \end{tabular}
    %\end{scriptsize}
    \caption{Knowledge graph topics, entities and relations.       \label{table:kg_topics}}
    %\vspace{-0.15in}
\end{wraptable}
 One limitation of the KG RGs include having to select ``interesting'' relations and write templates by hand. Another is the fact that most responses are
individual fact turns, with TV shows as the only RG which currently has multi-turn  responses based on a single relation. In the future, we plan to use more
relations to generate new responses, expand to more topics, and integrate new databases, as well as explore neural generators for KGs.

\subsection{Centering RGs}
\label{sec:centering-rg}

Centering RGs are topical  retrieval-based generators where fun facts relevant to the current context are retrieved
according to their indexed discourse entities and concepts. We significantly expanded these from SGC3 with more supported topics, entities and concepts. Almost every topic in Athena has a corresponding Center RG. These Center RGs can interweave with both KG RGs and Flow RGs.
%\begin{wrapfigure}{r}{3.2in}
\begin{figure*}[b!ht]
    %\vspace{-0.15in}
    %\begin{scriptsize}
    \def\arraystretch{1.1}
    \small
    \begin{tabular}{@{} p{1.05in}|p{4.3in} @{}} \toprule
         Animals Flow-RG &   I'm super curious to hear your thoughts. If you could own any wild animal as a pet, which animal would it be?		\\
        USER & 	hedgehog\\
        Animals Flow-RG  & That's an interesting answer! I would choose to own a white tiger.  First of all, tigers are awesome, and white tigers in particular look very cool. Anyways,  So, what's your favorite animal?  \\
        USER & hedgehog					 \\
        Animals Flow-RG & hedgehogs are a good choice. What about hedgehogs makes them your favorite? \\
        USER &  cause they're adorable and they're tiny	 \\
       Animals Center-RG    &  I heard this trivia fact about hedgehogs. Hedgehogs are thought to be one of the oldest mammals on earth!\\
        USER & really	\\	
      Animals Center-RG  & A group of hedgehogs is called an array, and a young hedgehog is called a hoglet, a piglet or a pup. \\
        USER & no	\\
       %FLOW & Horses are so cool, and have such gorgeous manes! What do you like about them? 	\\		USER & 	they're fast	\\
% 		ATHENA &	If I had a horse as a pet, I would brush it's mane everyday! Do you have a favorite horse? 	
        \hline
    \end{tabular}
    %\end{scriptsize}
    \caption{Conversation showing Animals Center RG interweaving  with  Animals Flow-RG 
    \label{fig:animals-dialogue}}
    %\vspace{-0.15in}
    \end{figure*}
%\end{wrapfigure}

 An example of the Animals  Flow-RG interweaving with the Animals Center RG is in Figure~\ref{fig:animals-dialogue}. During SGC4,  we created a vastly expanded Sports Center RG with over 4,500 fun facts across many different sports. We also greatly increased the fun facts in our Animals Center RG since animals is such a popular topic. We created a Food topic, and added a lot of fun facts about chefs and foods to the Food Center RG. We also greatly expanded our Center RGs for Comic Books and Board Games. 

\subsection{Neural Response Generators}
\label{nrg-sec}

%\subsubsection{Amazon’s Topical Neural Response Generator}
%\label{tc-nrg-sec}
We developed and deployed in testing two different types of Neural Response Generators (NRGs). The Discourse-Driven NRG 
discussed in Section~\ref{dd-nrg-sec} is
trained End-to-End using GPT-2 with fine-tuning using
Topical Chat. The Video Games meaning-to-text NRG is trained from a corpus of meaning representation and reference utterance pairs \cite{juraska2019viggo}. Neither of these NRGs ran in our A branch because either they were too slow, or they were not deemed to be reliable in terms of their output. 
We are currently exploring methods to further control their output.
\subsubsection{Discourse-Driven-NRG}
\label{dd-nrg-sec}

\begin{wraptable}{r}{0.55\linewidth}
    \small
   	\centering
   	\def\arraystretch{1.15}
    \begin{tabular}{p{0.18\linewidth} p{0.66\linewidth}}
    \hline
         {\bf Context} &  \textbf{User 1:} oh awesome. the movie paid the rock 20 \$ million, that is a lot of money \\
         & \textbf{User 2:} wow that is and the total budget was \$ 91 so he got paid a good chunk \\
         & \textbf{User 1:} yeah that is true. do you use your phone much?\\
         \hline
         {\bf Dialogue Acts} & backchannel, yes-no-question \\
         \hline
         {\bf Knowledge} & unicef will donate a day of clean water to a child in need for every minute you do not use your cell phone. \\
         \hline
         \textbf{Response} & yeah. did you know that unicef will donate a day of clean water to a child in need for every minute you do not use your phone? \\
         \hline
    \end{tabular}
     \caption{A training example from the Topical Chats corpus illustrating a knowledge-grounded dialogue plan.}
    \label{tab:dd-nrg-dataset-examples}
\end{wraptable}
The Discourse Driven Neural Response Generator (DD-NRG)  is a neural response generator that generates utterances grounded on textual knowledge and a dialogue plan \cite{tosh2020,rajasekaran2020}. The RG builds on the Policy Driven Neural Response Generation (PD-NRG) approach, where a neural language model is trained to generate utterances whose surface form is guided by dialogue acts and knowledge provided in the input \cite{hedayatnia2020policy}. We use a more discriminative set of dialogue acts, and our own knowledge selected from  possible responses created by Athena's Center RG for each topic, providing a set of curated fun facts that were specifically retrieved to match the dialogue context. See Section~\ref{sec:centering-rg}.   The dialogue act schema used in the PD-NRG approach contains only 11 coarse-grained tags such as \textit{Feedback}, \textit{Statement} based on the dialogue act tagging scheme defined in \cite{mezza2018iso}.  This scheme does not distinguish dialogue acts such as positive and negative answers, opinion statements, and agreement and disagreements.  Our hypothesis is that conditioning response generation on a finer-grained dialogue act scheme will provide better control of the structure of the utterance. We collapse the Switchboard-DAMSL tagset into a reduced set of semantically similar tags and eliminate tags targeted only at spoken conversation \cite{kumar2017dialogue,Shriberg-etal:1998,stolcke2000dialogue,jurafsky1997switchboard}. We then train a BiLSTM-CRF tagger on the labelled Switchboard corpus. 

\noindent{\bf Training Data.}
We start with the Topical Chat  corpus of  crowdsourced human-human conversations grounded on external knowledge \cite{gopalakrishnan2019topical}. Topical Chat is one of the largest corpora available for open-domain dialogue. To create Topical Chat,  pairs of crowd-workers were given reading sets  of news article snippets, factoids and Wikipedia article sections, and then were instructed to carry out typed conversations grounded on the information provided in the reading sets.  Following the same methodology used by PD-NRG, we augment the Topical Chat corpus by annotating all turns with our trained Switchboard-DAMSL tagger.   
Each training instance is comprised of past context turns, the response, the dialogue acts used in the response, and an optional knowledge sentence. A sample training instance is shown in Table~\ref{tab:dd-nrg-dataset-examples}.

\noindent{\bf Model.} The Discourse Driven NRG is based on the GPT-2  architecture and uses the $\sim$377M parameter \texttt{gpt2-medium} pretrained weights provided by OpenAI \cite{radford2019language}. The model is fine-tuned on the dialogue-act tagged Topical Chats corpus on the tasks of language modeling and next utterance prediction, following the same approach used for the TransferTransfo model \cite{wolf2019transfertransfo}. The utterances are generated using nucleus sampling \cite{holtzman2019curious} which leads to greater diversity of responses, but causes the generative process to become non-deterministic. An early version of this NRG participated in the DSTC9 Shared Task \cite{gunasekara2020overview,tosh2020,rajasekaran2020}.

The inputs to the model are a set of dialogue acts (the policy) and knowledge, represented as text. We are currently 
exploring the degree of control we can exercise over the outputs depending on the knowledge and the policy. We have run experiments with  the policies in Table~\ref{tab:dd-nrg-policy-examples}; an example response is provided. We plan to conduct further experiments conditioning the policy on the user's dialogue act and the conversation history.

\begin{table*}
\centering
\def\arraystretch{1.15}
\begin{tabular}{|p{1.6in}| p{3.40in}|}
\hline
\multicolumn{2}{|p{5.0in}|}{\bf \centering Dialogue Context}  \\ \hline
    \multicolumn{2}{p{5.0in}}{ \textbf{Athena:} Yeah, I get that. <prosody rate='90\%'>I would want to be able to rap, because I cannot do it now, and the speed that rappers sing is incredible. 
        Also their ability to come up with raps and rhymes on the spot is something I wish I could do.</prosody> Anyway, just my opinion. Which musician do you want to discuss? }\\
  \multicolumn{2}{p{5.0in}}{   \textbf{User:} here's what i want to talk about i wanna talk about chess and the movie queen's gambit did you ever see that }\\
  \multicolumn{2}{p{5.0in}}{   \textbf{Athena:} Sure, I'd love to talk, we can talk about anything you want. Hhmm, anyways,  I love talking about movies! I would consider Gambit if you want to watch a heist film movie. }\\
  \multicolumn{2}{p{5.0in}}{   \textbf{User:} well this movie is queen's gambit and one best who is nominated for best picture} \\ 
     \hline %\hline
     \textbf{Dialogue Acts} & \textbf{Possible Responses}  \\
     \hline
     backchannel, statement-non-opinion, yes-no-question & yeah i think so. i think i have seen the movie, what about you?\\
     \hline
     backchannel, statement-opinion, wh-question & yeah i heard that. what do you think about the movie?\\
     \hline
     backchannel-in-question-form, statement-non-opinion, yes-no-question & yeah that's why i love it. do you like to watch movies?\\
     \hline
     agree, statement-opinion, yes-no-question & yeah that is a great movie, i liked it. do you like king of hill?\\
     \hline
     appreciation, statement-non-opinion, open-question & yeah i like that one too. did you know that you can buy movies on amazon.\\
     \hline
     backchannel, statement-opinion, open-question & yeah i think that's cool. do you watch television much?\\
     \hline
\end{tabular}
\caption{Example responses produced by different policies}
\label{tab:dd-nrg-policy-examples}
\vspace{-.2in}
\end{table*}
%\noindent{\bf Current Policies.}

\noindent{\bf Qualitative Evaluation.}
The Topical Chat corpus is a written chat corpus between two humans while the DD-NRG model is intended to operate in a real-time spoken conversation. We therefore   set up an evaluation task where human raters were asked to examine Athena's response in a particular dialogue context, and then annotate DD-NRG's response for the following attributes:

    \noindent \textbf{Makes Sense:} Is the DD-NRG response sensible and not just gibberish? \\ 
    \noindent \textbf{Coherent to Dialogue Context:} Is the DD-NRG response consistent with the previous dialogue context and user's current utterance. This is in line with the competition's goal of maintaining a coherent conversation with the user. \\
    \noindent \textbf{Takes Conversation Forward:} Does the DD-NRG response help keep the conversation going or makes it end abruptly. We hypothesize that ending the response with a question would help the DD-NRG maintain the flow of the conversation. \\
    \noindent \textbf{Better than current Athena response:} Is the DD-NRG response better than Athena's current response. If the response is similar or better, it's marked as positive, indicating that the DD-NRG response can match the quality of other hand crafted responses from Athena.

The first example in  Table~\ref{tab:dd-nrg-annotation-examples} has a  DD-NRG response  tagged as \textit{Makes Sense} and \textit{Takes Conversation Forward}, since it asks for the cat's name, but \textit{Not Coherent to Context} as it assumes that the user is talking about their pet cat. In this case, Athena's current response is better as it gives a general fact about cats. In the second example, the DD-NRG does a better job of acknowledging the user's question about Athena having any pets, and then asking a related follow-up question about dogs. This example was annotated as positive for all four metrics.

%\newne

\begin{table*}[b!ht]
\centering
\def\arraystretch{1.15}
\begin{tabular}{p{.6in} p{4.6in}}
\hline
     \textbf{Context}
     & \textbf{Athena:} No worries. I was wondering, do you think it's easier to break the ice with a human, or an AI like me? \\
     & \textbf{User:} human\\
     & \textbf{Athena:} That makes sense. After having so many wonderful conversations with people like you, I can only look forward to getting better in the future. I've been trying to learn more about all kinds of cool animals in the wild, I wish I could see them all. One of my favorites is the koala bear.\\
     & \textbf{User:} my favorite the cheetos actually i like all big cats\\
     \hline
     \textbf{Athena} & Here's a fact you may know about cats! While cats can distinguish their own name, they don't necessarily feel obligated to respond. \\
     \hline
     \hline
     \textbf{DD-NRG} & yeah i agree. do you know what the name of the cat is? \\
     \hline
     \textbf{Context}
     & \textbf{User:} how old are you\\
     & \textbf{Athena:} I am a product of many people whose experience adds up to thousands of years. So, in a way I am thousands, maybe millions, of years old. Anyways, Hmmm, I have to keep thinking of ways to answer all the great questions I get asked. I appreciate you wanting to learn more about me, and I definitely want to keep chatting.\\
     & \textbf{User:} do you have any pets\\
     \hline
     \textbf{Athena} & If I was to have a pet it would be a cat, I've seen videos online of them, and they look cuddly. Do you have a pet? If you don't mind me asking. \\
     \hline
     \textbf{DD-NRG} & yeah i'm a dog person. do you like dogs? \\
     \hline
\end{tabular}
  
\caption{Evaluation examples based on Athena conversations on Animals}
\label{tab:dd-nrg-annotation-examples}
\end{table*}

\textbf{Results.} We annotated 1,020 examples across 11 topics for the evaluation. The topics were a mix of topics in Topical Chat such as Books, Sports, Music, and Movies as well as some Athena-specific topics like Animals, Astronomy, Harry Potter and Video Games. Overall, we observed that \textbf{69.31\%} of DD-NRG responses made sense while \textbf{72.74\%} of the responses took the conversation forward by asking relevant follow-up questions. The overall coherence to conversation was \textbf{53.82\%} but jumped up to \textbf{63.85\%} if we only consider the topics available in Topical Chat. Thus, as expected, DD-NRG  conditions better on contexts similar to its training data. Finally, we observed that \textbf{32.35\%} of the responses were either better or comparable to Athena's current responses. While this metric may be low, our aim is to use  DD-NRG  to augment Athena's response generation capabilities specifically in situations where the system is unable to provide a good response. 

\subsubsection{Neural Response Generation from Meaning Representations}
% Owner: Jurik, reviewer comment was that we should call it this

Another way we make use of neural models in Athena is for  generating responses from meaning representations \cite{reedetal20,oraby2019curate,harrison2019maximizing,juraska2018deep,Juraskaetal18}.   This approach relies on the content and the dialogue act (DA) to be provided as input, but offers more control over the generated response in return. It is therefore compatible with the Flow-RG framework, and can be used to generate whole responses, as well as just their segments (combined with template-based segments), at any point in the flow-graph. Flow-RG prepares the input meaning representation (MR) for the model -- essentially a dictionary of attributes and their values, which constitute the desired content to mention, along with the DA type indication -- at any given node in the flow graph, depending on the context and the user model.

This hybrid neural-and-template design drastically reduces the amount of effort put into writing response templates for variable content (typically content that can be looked up in a knowledge base), yet it takes advantage of a relatively small number of handcrafted templates to achieve natural and context-aware acknowledgments and transitions, which the data-to-text model is not trained to produce. The handcrafted templates are mostly short and do not contain any slots to fill with information. Note that a response being composed of a mixture of neural-generated and template-based segments does not change anything about how Flow-RG ultimately combines them into a final response.

\begin{wraptable}{r}{0.55\linewidth}
    \small
   	\centering
   	\vspace{-.1in}
   	\def\arraystretch{1.5}
    \begin{tabular}{p{0.94\linewidth}}
    	\hline
    	\rowcolor{light-gray}
    	\emph{confirm}(\textsc{name} [\textbf{Hellblade: Senua's Sacrifice}], \textsc{release\_year} [\textbf{2017}], \textsc{developer} [\textbf{Ninja Theory}]) \\
        \hline
    	Oh, do you mean the \textbf{2017} game from \textbf{Ninja Theory}, \textbf{Hellblade: Senua's Sacrifice}? \\
  %      \hline
   % 	\rowcolor{light-gray}
   % 	\emph{give\_opinion}(\textsc{name} [\textbf{SpellForce 3}], \textsc{rating} [\textbf{poor}], \textsc{genres} [\textbf{real-time strategy, role-playing}], \textsc{play\-er\_per\-spec\-tive} [\textbf{bird view}]) \\
  %      \hline
  %  	I think that \textbf{SpellForce 3} is \textbf{one of the worst games} I've ever played. Trying to combine the \textbf{real-time strategy} and \textbf{role-playing} genres just doesn't work, and the \textbf{bird's eye view} makes it near impossible to play. \\
    %    \hline
    %	\rowcolor{light-gray}
    %	\emph{verify\_attribute}(\textsc{name} [\textbf{Little Big Adventure}], \textsc{rating} [\textbf{average}], \textsc{has\_multiplayer} [\textbf{no}], \textsc{platforms} [\textbf{PlayStation}]) \\
  %      \hline
   % 	I recall that you were \textbf{not that fond} of \textbf{Little Big Adventure}. Does \textbf{single-player} gaming on the \textbf{PlayStation} quickly get boring for you? \\
        \hline
    \end{tabular}
%  	\vspace{-0.1cm}
	\caption{Examples of four dialogue acts and their  MRs (gray rows) and the corresponding reference utterances (with slot mentions in bold) in the ViGGO dataset. DA types are indicated in italics at the beginning of the MRs.}
    \label{tab:viggo-dataset-examples}
%  	\vspace{-0.3cm}
\end{wraptable}
%\noindent{\bf Training Data.}
\noindent{\bf Training Data.}
Making use of the ViGGO dataset~\cite{juraska2019viggo}, we integrate data-to-text generation into our flow-based video game RG. Since ViGGO was created with chit-chat conversations about video games in mind -- rather than task-oriented conversations, which is typically the case with data-to-text datasets -- training an NLG model for a socialbot on this dataset seemed like a natural application, without the need to augment or annotate the dataset. The flow graph of the RG is designed around the 9 DAs defined in the dataset, such as \emph{verify attribute}, \emph{request explanation}, or \emph{recommend}. Most of the DAs are compatible with many different combinations of content slots, corresponding to 14 different video game attributes, resulting in hundreds of types of utterances a model trained on this dataset can produce. Table~\ref{tab:viggo-dataset-examples} shows examples of three DAs with various combinations of slots.

\noindent{\bf Model.}
In our experiments, we used the smallest variants of the pretrained encoder-decoder language models T5 and BART~\cite{raffel2020exploring,lewis2020bart}, which we fine-tuned on the 5K training examples of the ViGGO dataset. With $\sim$60M parameters, the T5-small is faster for real-time inference than the $\sim$ 139M-parameter BART-base, while their performance in terms of coherence and semantic accuracy is comparable. We therefore deploy T5-small in the production system, nevertheless, to keep Athena's response latency within reasonable bounds, we restrict the model to greedy decoding instead of beam search, which results in a negative impact on the quality of the generated responses.

% TODO(Jurik): Add a table with automatic metric results, as well as SER, for these two models.

\begin{table}
    \small
   	\centering
   	\def\arraystretch{1.15}
    \begin{tabular}{p{0.08\linewidth} >{\raggedright\arraybackslash}p{0.86\linewidth}}
		\hline
    	ATHENA:  & \dots Well, one of my all-time favorites is Control. I have already completed it twice! Have you played it by any chance? \\
    	\hline
    	\rowcolor{light-gray}
    	USER:  & no not yet \\
    	\hline
    	ATHENA:  & Obviously. There are just too many fun games for one person to play in a single lifetime. At any rate, \textcolor{burntorange}{\textbf{Control} is an \textbf{excellent} \textbf{third-person} \textbf{action-adventure} \textbf{shooter} with \textbf{multiplayer}. It was released in \textbf{2019} by \textbf{Remedy Entertainment} for \textbf{PC}, \textbf{PlayStation}, and \textbf{Xbox}.} Do you think you would potentially like it? \\
    	& \textcolor{darkgray}{\emph{inform}(name[\textbf{Control}], release\_year[\textbf{2019}], developer[\textbf{Remedy Entertainment}], rating[\textbf{excellent}], genres[\textbf{shooter}, \textbf{action-adventure}], player\_perspective[\textbf{third person}], has\_multiplayer[\textbf{yes}], platforms[\textbf{PC}, \textbf{PlayStation}, \textbf{Xbox}])} \\
    	\hline
    	\rowcolor{light-gray}
    	USER:  & absolutely i was actually hoping to play it sometime soon \\
    	\hline
    	ATHENA:  & Alright! \textcolor{burntorange}{Have you played any \textbf{adventure} games by \textbf{Remedy Entertainment}, like \textbf{Alan Wake}?} \\
    	& \textcolor{darkgray}{\emph{suggest}(name[\textbf{Alan Wake}], developer[\textbf{Remedy Entertainment}], genres[\textbf{adventure}])} \\
    	\hline
    	\rowcolor{light-gray}
    	USER:  & yes i did i loved the dark atmosphere in alan wake \\
    	\hline
    	ATHENA:  & Yeah! I remember playing Alan Wake some time ago, and it wasn't bad at all. \textcolor{burntorange}{Do you like \textbf{multiplayer} games?} \\
    	& \textcolor{darkgray}{\emph{request\_attribute}(has\_multiplayer[])} \\
    	\hline
    \end{tabular}
 	\vspace{0.1cm}
	\caption{An excerpt of a conversation with Athena about video games, exhibiting a smooth interweaving of neural and template-based responses of the video game RG. Response segments generated by the neural model are highlighted in orange. For each such segment, the MR constructed by the RG as input for the model is shown below the response.}
    \label{tab:video-game-conversation-example}
 	\vspace{-0.1cm}
\end{table}

\noindent{\bf Qualitative Evaluation.}
The example in Table~\ref{tab:video-game-conversation-example} demonstrates three subsequent turns in which Athena makes use of the neural data-to-text generator. The response in the third row contains the most complex segment, involving 8 different slots, two of which have values listing multiple elements. The neural model did a great job chaining the mentions of three of the slots in a natural order in ``excellent third-person action-adventure shooter'', while also reordering the individual elements of the \emph{genres} slot and omitting the word ``game'' instead of generating something less natural like ``shooter action-adventure game''. The latter would have likely been the result if a general template was used instead, in which the \emph{genres} slot was filled by simply concatenating the elements of its value. Another detail in the generated segment worth noting is the indefinite article before ``excellent'', which is something that needs to be explicitly accounted for, along with singular vs. plural noun forms, when using templates. Although the other two generated response segments in the example are simpler, it is clear the neural model produces coherent and semantically correct text that can be seamlessly combined with one or more templates into a full response.

As we can see in the first and last turn of the conversation excerpt, there is definitely scope for expanding the coverage of the neural generator to more types of responses. ViGGO, however, does not contain examples of those response types that the model could learn from, so we handle them using templates.

\section{User Model}
\label{sec:user_model}
% Owner: Kevin

Creating a user model, and modifying the dialogue system's responses
based on that user model, should increase the user's feelings of agency, and create  more engaging conversations 
\cite{komatani2005user, o2009using}. We build a user
model incrementally over multiple conversations with each user. 

The user model relies on the NLU pipeline, several handcrafted regular expressions, and annotated data. The user model tracks general information across conversations, such as the user's name, whether or not they've self-identified as a youth, and their interests. We also track topic-specific information, such as their pet's names, their weekend hobbies, and their favorite dinosaur.

We primarily use the user model when determining which topics to present the user with, either in a menu of options or by explicitly taking the initiative. The user model
promotes topics that it believes users are most likely interested in hearing 
about without needing to explicitly ask for their input as well as, within a topic, tailoring 
our responses based on how the user responds. 

To increase the user's feelings of agency \cite{harrell2009agency},  we adapt Athena's responses  to the values stored in the user model. For example, if Athena is taking the initiative when picking the next topic and earlier the user said \textit{I like eating pasta}, Athena will initiate the food topic, e.g., \textit{I remember your interest in food, so why don't we talk about food! I've become really interested in food lately! It's really cool learning about all the different cuisines. I'm curious, would you go so far as to say you're a big foodie?}. Here, Athena shows that she remembers the user, and initiates  a new topic that is likely to be relevant to the user's interests.

We further use this information to better handle repeat user conversation. For example, we use the user's name to personalize our greeting, to avoid topics that we've already explored, and remember their previous preferences. In live A/B testing we have seen a statistically significant improvement in user rating when comparing our personalized topic selection strategy against our default topic selection strategy.
 
 \begin{figure}[h]
 \vspace{.2in}
  \centering
  \includegraphics[width=\textwidth]{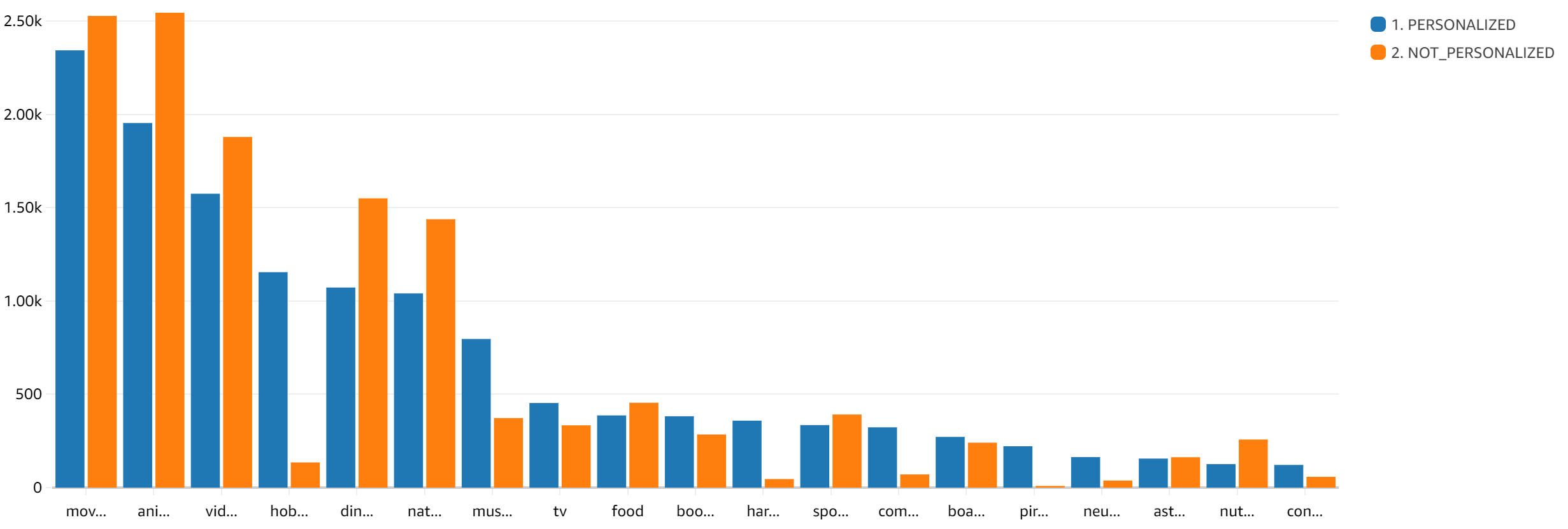}
  \caption{A comparison of the distribution of turns spent in each topic in  our personalized topic selection (\textit{Version A} in blue) and default selection (\textit{Version B}) in orange.
  \label{fig:ab_topic_turn_dist}}
  %\vspace{-.1in}
\end{figure}
\noindent{\bf Personalized Topic Selection.}
Using the personalized selection strategy detailed in Section~\ref{sec:user_model}, we can see a difference in the number of turns spent in each topic. Figure~\ref{fig:ab_topic_turn_dist} shows a comparison between the personalized topic selection strategy (a\_version) and the default topic selection strategy (b\_version). One notable example is our Hobbies topic, which in the personalized case is triggered significantly more than in the random selection case, primarily due to the weekend hobbies stored in the user model during the introduction (Section~\ref{sec:intro_rg}).

\section{Evaluation Analysis and Results}
\label{eval-sec}

% \subsection{Dialogue Manager Error Analysis}
% % Owner: Omkar
% The goal of this study is to identify Athena's mistakes to make better dialogue management decisions. We hand annotated Athena's responses to identify 204 mistakes most of which (\textbf{64.71\%}) were about Athena ignoring the user. We define "ignores user" as either the system does not acknowledge the content present in user's utterance or does not follow or acknowledge action directives.

%\subsection{User Ratings}
% Owner: Omkar

% \begin{figure}[tb]
%   \centering
%   \vspace{-.1in}
%   \includegraphics[width=\textwidth]{figures/Ratings-by-Topic-Jan1-to-June16.png}
%   \caption{Distribution of Ratings for each Topic for the period from January 1st to June 16th}
%   \label{fig:ratings-by-topic-distrib}
%   \vspace{-.15in}
% \end{figure}

Evaluation of interactive dialogue is extremely challenging \cite{WLKA97,WKL00,Walkeretal02a}. In the context of Athena,
Amazon asks users for their ratings after every conversation, on a scale of 1 $\ldots$ 5. It is obvious that a user's interactions with different response generators (RGs) and topics affect their
views of the system and therefore their ratings. However, ratings are only collected at the end of the dialogue, and only about 20\% of users actually provide dialogue ratings. 

\begin{figure}[htb]
  \centering
  \vspace{-.1in}
  \includegraphics[width=\textwidth]{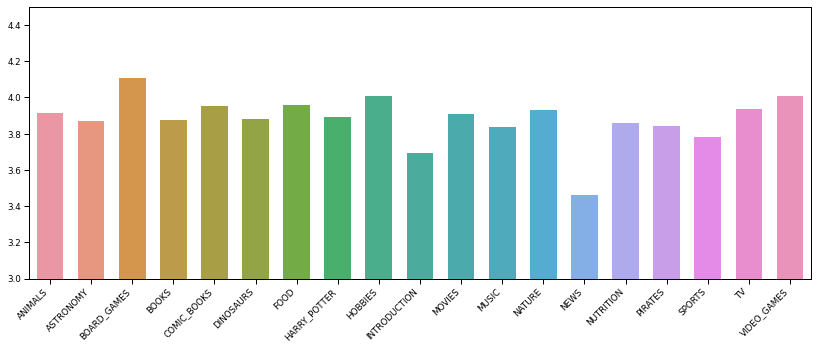}
  \caption{Average Rating per Topic in June}
  \label{fig:average-rating-by-topic}
  %\vspace{-.15in}
\end{figure}

We calculate the performance  of the Response Generators for each topic by assuming that if an RG is used for at least 3 turns in a dialogue, then its pool of ratings include the rating for that dialogue. Using this simplifying assumption, ratings for each RG are  shown in Figure~\ref{fig:average-rating-by-topic}.
It is easy to see by examining the blue and orange bars indicating ratings of 4 and 5 respectively, that the strongest topics include animals, comic books, harry potter, hobbies, and video games, but that only a few topics are actually performing poorly, such as dinosaurs, news and sports.

\begin{figure}[h!tb]
  \centering
  \vspace{-.1in}
  \includegraphics[width=\textwidth]{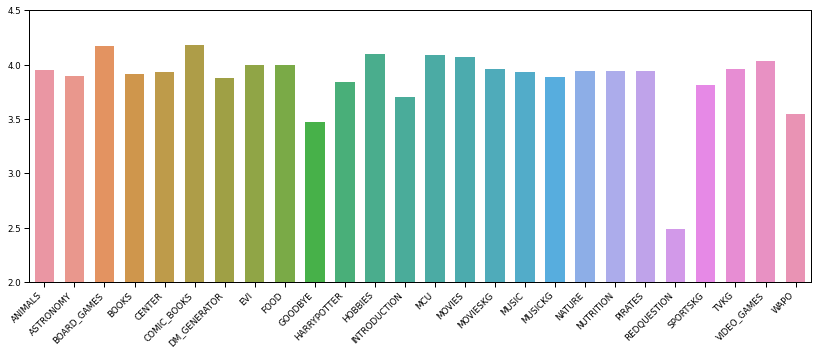}
  \caption{Ratings by response generator during June}
  \label{fig:average-rating-per-RG}
  \vspace{-.05in}
\end{figure}

\begin{figure}[htb]
  \centering
  \includegraphics[width=\textwidth]{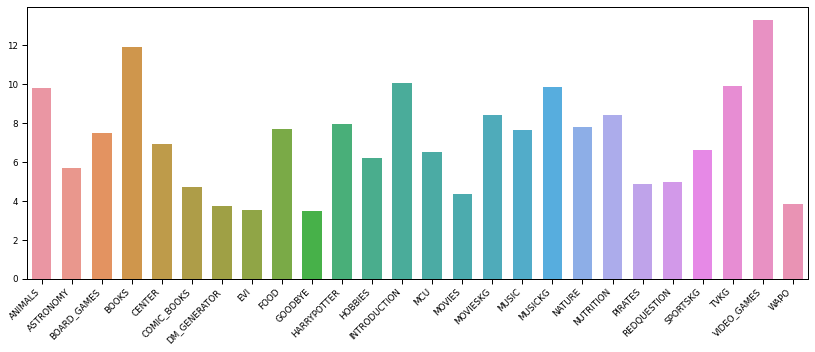}
  \caption{Turn Length per Response Generator in June}
  \label{fig:turn-length-per-RG}
  %\vspace{-.15in}
\end{figure}

In order to calculate the average
rating per specific RGs or specific topics, we assume that conversations that contain turns with a specific RG
or topic, allow us to estimate how that RG or topic contributes
the user rating. 
In order to examine Athena's performance in terms of user ratings, we first filter rated dialogues by taking into account that some users seem to accidentally get into an Alexa Prize conversation and leave immediately. Also sometimes, Athena introduces a topic that the user isn't interested in. To examine the effect of a topic or RG on ratings, we exclude  subdialogues  that have less than 3 turns. In Figure~\ref{fig:average-rating-per-RG}, we 
can see that RGs such as red question and goodbye are very poorly rated. This is because
red question responses tend to occur with  adversarial users. 
%and also, taking 
%more than two turns to leave the conversation will %lead to frustration and a lower rating. 
Board games and comic books have the highest conversation ratings. These RGs are two of the 
smaller Flow-RGs which have clear entities, allowing the Centering RG to interject 
interesting fun facts. They are also topics that users in general find interesting and 
fun to talk about. We also report the average turn length for all RGs in 
Table~\ref{fig:turn-length-per-RG}. The RG video games has the longest average turn count, 
with 13.3 average turns per conversation. Video 
games is a fact-based Flow-RG and is able to talk about multiple video games in a single 
conversation. Other RGs that gather information from a database, such as books, musickg, 
movieskg and tvkg also have longer turn averages. 

In Figure~\ref{fig:average-rating-by-topic}, we can see that, again, board games has the highest
average rating. Hobbies is the second highest for topics. The hobbies topic attempts to first 
discuss hobbies that interest the user. Since hobbies is such a broad topic, most people have 
a hobby and are therefore likely to find something engaging about this topic. 

\begin{figure}[htb]
  \centering
  \vspace{-.1in}
  \includegraphics[width=\textwidth]{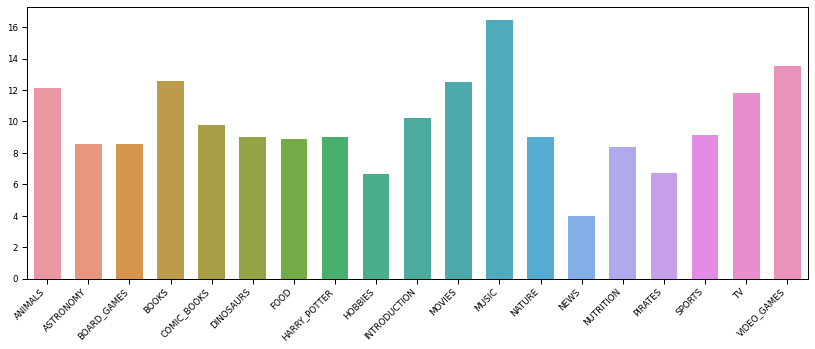}
  \caption{Average Turn length per topic}
  \label{fig:turn-length-by-topic}
  \vspace{-.15in}
\end{figure}

Figure~\ref{fig:turn-length-by-topic} shows the average turn length per topic. Here, music
has the highest number of average turns, with 16.4, almost 3 turns more than the second highest. 
Music has three very different RGs which the user can interact with. First, the Centering
 RG provides 31 interesting facts that Athena can respond with, allowing for 
hand-curated facts on popular musicians and bands. The MusicKG, can have long conversations
on musicians and bands. And, finally, Music Flow-RG, which, instead of focusing on entities, converses on 
the user's opinions on music, both on listening to music and playing instruments. This allows for the user
to talk about a single topic without Athena running out of engaging responses.

\section{Conclusion}
Here we describe a conversational agent for spoken dialogue named Athena,
which competed as an Alexa Prize Socialbot in 2020/2021. We interpret user input 
using various methods, such as coreference resolution, NER/NEL and user modeling. We 
tune our responses based on these features to create responses which fit the 
user's interests. 

One of the strengths of  Athena is its ability to incorporate responses from a wide variety of resources, such as real-time KG walks, as well as a modular conversation flow framework, that allows RGs to be interweaved. We greatly improved our NER/NEL system that incorporates a large knowledge base of entities and methods for generating a huge corpus of synthetic training examples, as well as utilizing annotated data from real user conversations. We developed a discourse model and a coreference resolution module which integrates our use of the Wikidata KG across multiple response generators, allowing them to maintain context and understand what has been discussed. We also  experimented with several neural methods to generate responses and analyzed the quality of these outputs.  Our analysis of  Athena's conversations collected during the semi-finals phase suggests that Athena 2.0 is greatly improved over Athena 1.0, with many of our response generators engaging the users to a greater depth. But obviously, with a challenge like the Alexa Prize Challenge, there is always future work to be done. In future work, we hope to further improve our representation of context, and further explore the use of neural response generation in order to reduce the amount of handcrafting in conversational agents.

\subsubsection*{Acknowledgments}
Thank you to Navya Rao, Venkatesh Nagubandi, Diego Pedro, Jasiel Rivera-Trinadad, Christian Benavidez,
Meltem Ozcan, Stephanie Rich, Yaqing Cao, James Graupera, Colin Harmon and Aditya Tarde
for helping with fun facts and Wikidata queries.
Thanks also to Kate Bland, Savanna Stiff, Shui Hu, Anna Gottardi and other members of the Alexa Team at Amazon who gave us feedback and technical assistance during the year.
%Use unnumbered third level headings for the acknowledgments. All
%acknowledgments go at the end of the paper. Do not include
%acknowledgments in the anonymized submission, only in the final paper.

%\section*{References}
%References follow the acknowledgments. Use unnumbered first-level
%heading for the references. Any choice of citation style is acceptable
%as long as you are consistent. It is permissible to reduce the font
%size to \verb+small+ (9 point) when listing the references. {\bf
%  Remember that you can use a ninth page as long as it contains
%  \emph{only} cited references.}

%\small
%\footnotesize
\bibliographystyle{plain}
%\nopagebreak
%\bibliography{athena,nl}
%\bibliography{nl}
%\bibliography{athena} % I copied them over.. moving on. 

\end{document}